%% file: main.tex
\documentclass{article}

\usepackage{PRIMEarxiv}

\usepackage[utf8]{inputenc} 
\usepackage[T1]{fontenc}    
\usepackage{hyperref}       
\usepackage{url}            
\usepackage{booktabs}       
\usepackage{amsfonts}       
\usepackage{nicefrac}       
\usepackage{microtype}      
\usepackage{lipsum}
\usepackage{amsmath}
\usepackage{fancyhdr}       
\usepackage{graphicx}       
\graphicspath{{media/}}     
\usepackage{amssymb}
\usepackage{multirow}
\usepackage{adjustbox}
\usepackage{listings}
\usepackage{siunitx}
\usepackage{xspace}
\usepackage{subcaption}

\usepackage[colorinlistoftodos]{todonotes}
\usepackage{xcolor}
\newcommand{\docllm}[0]{\texttt{DocLLM}\xspace}

\title{DocLLM: A layout-aware generative language model for multimodal document understanding
}

\author{
  Dongsheng Wang$^*$, Natraj Raman$^*$, Mathieu Sibue$^*$ \\ \textbf{Zhiqiang Ma, Petr Babkin, Simerjot Kaur, Yulong Pei, Armineh Nourbakhsh, Xiaomo Liu}  \\
  JPMorgan AI Research \\
  \texttt{\{first.last\}@jpmchase.com} \\
}

\begin{document}
\maketitle

\def\thefootnote{*}\footnotetext{These authors contributed equally to this work.}\def\thefootnote{\arabic{footnote}}

\input{sec_abstract}


\input{sec_intro}

\input{sec_relatedwork}

\section{DocLLM Framework}
\label{sec:pretrain_disentangled}

In this section, we discuss the architecture of \texttt{DocLLM} and outline the pre-training and instruction tuning procedures. Figure ~\ref{fig:modelarch} presents an overview of the model architecture.

\input{sec_pretrain}

\input{sec_instructtune}

\section{Experiments}
\label{sec:experiments}

\subsection{Datasets}
We gather data for pre-training from two primary sources: (1) IIT-CDIP Test Collection 1.0 \cite{lweis2006cdip} and (2) DocBank \cite{li2020docbank}. IIT-CDIP Test Collection 1.0 encompasses a vast repository of over 5 million documents, comprising more than 16 million document pages. This dataset is derived from documents related to legal proceedings against the tobacco industry during the 1990s. DocBank consists of 500K documents, each featuring distinct layouts and a single page per document. The relevant statistics for the datasets utilized in the pre-training are detailed in Table \ref{tab:pretrainingstatistics}. We obtain a collection of 16.7 million pages comprising a total of 3.8 billion tokens.

\begin{table}[]
\footnotesize
\centering
\caption{Pre-training dataset statistics.}
\label{tab:pretrainingstatistics}
\begin{tabular}{@{}lccccc@{}}
\toprule
& \textbf{No. of Docs} & \textbf{No. of Pages} & \textbf{No. of Total Tokens}   
\\ 
\midrule
\textbf{CDIP} & 5,092,636          & 16,293,353             & 3,637,551,478         \\
\textbf{DocBank} &    499,609                   & 499,609                   &  228,362,274    \\ 
\hline
\textbf{Total} &  5,592,245           & 16,792,962          & 3,865,913,752            \\ 
\bottomrule
\end{tabular}
\end{table}


We have introduced the datasets used to conduct instruction tuning on Section~\ref{sec:instruction_tuning}. These datasets encompass four common DocAI tasks: VQA, NLI, KIE, and CLS. Note that when a prompt includes a list of possible answers, we create multiple copies of the prompt with one possible answer assigned to each. We only perform this ``flattening'' 
operation in the training split of the dataset. Detailed statistics for these tasks are presented in Table~\ref{tb:instruct-stats}.


\begin{table}[]
\caption{\label{tb:instruct-stats}Instruction-tuning dataset statistics.
}
\small
\centering
\begin{tabular}{@{}lrr@{}}
\toprule
\textbf{Tasks}      & \textbf{No. of Training}    & \textbf{No. of Testing}  \\ \midrule
VQA    & 145,090  & 24,347 \\
NLI    & 104,360  & 12,720 \\
KIE     & 236,806 & 38,039 \\
CLS & 149,627  & 21,813 \\\hline
\textbf{Total} & 635,883   & 96,919 \\ \bottomrule
\end{tabular}
\end{table}

\subsection{Model Setup and Training Details}

Table \ref{tb:key_settings} provides key settings and hyperparameters for two variants of \docllm: \texttt{DocLLM-1B}, which is based on the Falcon-1B architecture \cite{penedo2023refinedweb}, and \texttt{DocLLM-7B}, which is based on the Llama2-7B architecture \cite{touvron2023llama}\footnote{Since Llama2 does not come with pre-trained weights at 1B parameters, we use the Falcon-1B architecture for the smaller version of \texttt{DocLLM}.}. \texttt{DocLLM-1B} is composed of 24 layers, each with 16 attention heads and a hidden size of 1,536. \texttt{DocLLM-7B} comprises 36 layers, 32 heads, and a hidden size of 4,096. Using pre-trained weights as the backbone for the text modality, we extend the Falcon-1B and Llama2-7B models by adding the disentangled attention and block infilling objective as described in Section \ref{sec:pretrain_disentangled}.

For \texttt{DocLLM-1B}, we use a pre-training learning rate of \num{2e-4} with 1,000 warmup steps, employing a cosine scheduler, and Adam optimizer \cite{KingmaB14} with $\beta_1=0.9, \beta_2=0.96$ and a weight decay of 0.1. For instruction tuning we use a learning rate of \num{1e-4} with 500 warmup steps and a cosine scheduler, and the same parameters for weight decay and Adam optimizer as the pre-training phase. The Adam epsilon is set to \num{1e-5}. We pre-train for one epoch, and instruct-tune for a total of 10 epochs. 

For \texttt{DocLLM-7B}, pre-training involves a learning rate of \num{3e-4} with 1,000 warmup steps and cosine scheduler, weight decay of 0.1, and Adam optimizer with $\beta_1=0.9, \beta_2=0.95$. Instruction tuning uses a learning rate of \num{1e-4} with 500 warmup steps and a cosine scheduler, weight decay of 0.1, and Adam optimizer with $\beta_1=0.9, \beta_2=0.95$. Adam epsilon is set at \num{1e-6}. We conduct one epoch of pre-training, followed by three epochs of instruct-tuning, considering available computing resources.

The maximum sequence length, or context length, is consistently set to 1,024 for both versions during the entire training process. The \texttt{DocLLM-7B} models are trained with 16-bit mixed precision on 8 24GB A10g GPUs using fully sharded data parallelism, implemented with the accelerate library.\footnote{\url{https://huggingface.co/docs/accelerate}} The \texttt{DocLLM-1B} model, on the other hand, is trained on a single 24GB A10g GPU.

\begin{table}[]
\centering
\footnotesize
\caption{\label{tb:key_settings} Model configuration and training hyperparameters setting for \texttt{DocLLM-1B} and \texttt{-7B}. }
\begin{tabular}{@{}l|cc|cc@{}}
\toprule
                & \multicolumn{2}{|c}{\textbf{DocLLM-1B}}     & \multicolumn{2}{|c}{\textbf{DocLLM-7B}}     \\ \midrule
Backbone          & \multicolumn{2}{|c}{Falcon-1B \cite{penedo2023refinedweb}}                     & \multicolumn{2}{|c}{Llama2-7B \cite{touvron2023llama}}                     \\
Layers          & \multicolumn{2}{|c}{24}                     & \multicolumn{2}{|c}{36}                     \\
Attention heads & \multicolumn{2}{|c}{16}                     & \multicolumn{2}{|c}{32}                     \\
Hidden size     & \multicolumn{2}{|c}{1536}                   & \multicolumn{2}{|c}{4096}                   \\
Precision       & \multicolumn{2}{|c}{bfloat16}               & \multicolumn{2}{|c}{bfloat16}               \\ 
Batch size       & \multicolumn{2}{|c}{2}               & \multicolumn{2}{|c}{5}               \\
Max context length      & \multicolumn{2}{|c}{1,024}               & \multicolumn{2}{|c}{1,024}               \\\midrule
\textbf{}  & \textbf{Pre-train} & \textbf{Instruct-tune} & \textbf{Pre-train} & \textbf{Instruct-tune} \\ \midrule
Learning rate   & \num{2e-4}              & \num{1e-4}                   & \num{3e-4}              & \num{1e-4}                   \\
Warmups         & 1000              & 500                    & 1000              & 500                    \\
Scheduler type  & cosine            & cosine                 & cosine            & cosine                 \\
Weight decay    & 0.1               & 0.1                    & 0.1               & 0.1                    \\
Adam $\beta$s           & (0.9, 0.96)       & (0.9,0.96)             & (0.9,0.95)        & (0.9,0.95)             \\ 

Adam epsilon & \num{1e-5} & \num{1e-5}& \num{1e-6}& \num{1e-6} \\ \bottomrule
\end{tabular}
\end{table}

\subsection{Downstream Evaluation}

\textbf{Experimental settings}. We investigate two experimental settings: 
\begin{itemize}
    \item \textbf{Same Datasets, Different Splits} (SDDS): Following previous work in VRDU \cite{lee2023pix2struct,davis2022dessurt,kim2022donut,tang2023udop,ye2023mplugdocowl,ye2023ureader}, we first evaluate \texttt{DocLLM} on the unseen test split (or dev split when test split is unavailable) of each of the 16 datasets composing the instruction-tuning data. The motivation behind this very typical setting is to check how \texttt{DocLLM} performs when tasks and domains supposedly stay the same from train to test. 
    \item \textbf{Same Tasks, Different Datasets} (STDD): Following \cite{wei2022flan1, chung2022flan2, dai2023instructblip, zhang2023llavar}, we also evaluate \texttt{DocLLM} on held-out datasets. More precisely, we instruction-tune the pre-trained checkpoint of \texttt{DocLLM} on prompts from 11 of the 16 datasets considered in SDDS, then evaluate \texttt{DocLLM} on the test split of the remaining three datasets. The rationale behind this evaluation setting is to assess the performance of \texttt{DocLLM} when tasks are unchanged but domains and layouts differ from train to test. We believe examining this setting in the DocAI field is relevant because industry use cases usually encountered in practice revolve around VQA, KIE, and CLS, while document characteristics tend to change more often in production. We specifically isolate DocVQA, KLC, and BizDocs for STDD evaluation in order to (1) exclude at least one dataset per task from SFT when possible, (2) leave enough datapoints per task in the training split of the instruction-tuning data, (3) avoid data leakage (certain datasets were obtained from the same sources), and (4) benchmark models on popular yet challenging datasets when possible. Due to the high cost of instruction-tuning, we were not able to run additional experiments with different held-out datasets.
\end{itemize}

\textbf{Baselines}. In SDDS and STDD, we benchmark \texttt{DocLLM} against comparably-sized and SOTA LLMs using Zero-Shot (ZS) prompts that contain the text extracted from each document using an OCR engine (excluding the spatial information) \cite{touvron2023llama,ouyang2022instructgpt}. In SDDS, we also report numbers from recent DocAI LLMs evaluated in a similar setting \cite{ye2023mplugdocowl,ye2023ureader}. As motivated in section \ref{sec:related_work}, we do not consider DocAI models that require task-specific fine-tuning \cite{kim2022donut,davis2022dessurt,lee2023pix2struct} and/or dataset specific prompts \cite{tang2023udop}, and instead focus on LLMs with out-of-the-box instruction following capability.

\textbf{Metrics}. Following previous work \cite{borchmann2021due,lee2023pix2struct,ye2023ureader,ye2023mplugdocowl}, we evaluate all VQA datasets using Average Normalized Levenshtein Similarity (ANLS) \cite{biten2019anls}, with the exception of VisualMRC, for which we use CIDEr \cite{vedantam2015cider} and WTQ, for which we use accuracy\footnote{This is done to remain consistent with the results reported by other SotA models.}. Performance on all CLS and NLI datasets is measured using accuracy. We evaluate all KIE datasets with the F1 score.


\textbf{Results.} In the SDDS setting, as shown in the Table \ref{tb:sdds_main}, we observe that \texttt{DocLLM-7B} excels in 12 out of 16 datasets, inclusively compared to ZS results of GPT4 and Llama2, and SDDS results of mPLUG-DocOwl and UReader. Among equivalent models (excluding GPT4), our model outperforms in 14 out of 16 datasets. Specifically, \texttt{DocLLM} demonstrates superior performance in layout-intensive tasks such as KIE and CLS. In VQA and NLI, its performance surpasses that of most multimodal language models, although it underperforms compared to GPT-4. GPT-4 outperforms \texttt{DocLLM} in VQA, possibly due to the higher complexity of reasoning and abstraction involved in VQA datasets compared to tasks like KIE or CLS. \texttt{DocLLM-1B} demonstrates performance close to that of our larger model, suggesting that the smaller model can derive significant benefits from the architecture of \texttt{DocLLM}. 

In the STDD setting, our model demonstrates superior performance compared to Llama2 across four out of five datasets, and achieves the best score overall for two of them (KIE task again). DocLLM also outperforms  mPLUG-DocOwl on DocVQA and both mPLUG-DocOwl and UReader on KLC, despite both baselines having been instruction-tuned on these datasets. However, it is important to note that classification accuracy is notably lower in our model. This discrepancy may stem from the fact that our model has been trained using only one classification dataset, limiting its ability to generalize effectively to new datasets.


\begin{table}[]
\centering
\caption{\label{tb:sdds_main}\centering Performance comparison in the SDDS setting against other multimodal and non-multimodal LLMs; non-multimodal LLMs are Zero-Shot (ZS) prompted while multimodal LLMs are instruction-tuned on the train split of the datasets considered. `-' marks not available.}
\centering
\footnotesize
\resizebox{\columnwidth}{!}{%
\begin{tabular}{@{}cc|cccc|cc@{}}
\toprule

\textbf{} &
  \multirow{3}{*}{\textbf{Dataset}} &
  \textbf{GPT-4+OCR} &
  \textbf{Llama2+OCR} &
  \textbf{mPLUG-DocOwl} &
  \textbf{UReader} &
  \textbf{DocLLM-1B} &
  \textbf{DocLLM-7B} \\
\multicolumn{1}{l}{} &
   &
  $\sim$1T (T) &
  7B (T) &
  $\sim$7B (T+V) &
  $\sim$7B (T+V) &
  1B (T+L) &
  7B (T+L) \\
                              &            & ZS    & ZS            & SDDS  & SDDS  & SDDS          & SDDS          \\ \midrule


\multirow{6}{*}{\textbf{VQA}} & DocVQA     & \textbf{82.8}         & 47.4          & 62.2           & 65.4           & 61.4          & \underline{69.5} \\
                              & WTQ \textit{(Accuracy)}       & \textbf{65.4}         & 25.0 & 26.9           & \underline{29.4}           & 21.9          & 27.1 \\
                              & VisualMRC \textit{(CIDEr)}  & \underline{255.1}        & 115.5         & 188.8          & 221.7          & 245.0         & \textbf{264.1} \\
                              & DUDE       & \textbf{54.6}         & 38.1 & -              & -              & 42.6             & \underline{47.2} \\
                              & BizDocs     & 76.4         & 48.8          & -              & -              & \underline{84.5}          & \textbf{86.7} \\ \midrule
\textbf{NLI}                  & TabFact   & \textbf{77.1}         & 48.2           & 60.2           & \underline{67.6}           & 58.0          & 66.4\\ \midrule
\multirow{9}{*}{\textbf{KIE}} & KLC        & 45.9         & 27.8          & 30.3           & 32.8           & \underline{58.9}          & \textbf{60.3} \\
                              & CORD       & 58.3         & 13.8           & -              & -              & \underline{66.9}          & \textbf{67.4} \\
                              & FUNSD      & 37.0         & 17.8          & -              & -              & \underline{48.2}          & \textbf{51.8} \\
                              & DeepForm   & 42.1         & 20.5          & 42.6           & 49.5           & \underline{71.3}          & \textbf{75.7} \\
                              & PWC        & 18.3         & 6.8           & -              & -              & \underline{25.7}          & \textbf{29.06} \\
                              & SROIE      & 90.6         & 56.4          & -              & -              & \underline{91.0}          & \textbf{91.9} \\
                              & VRDU a.-b. & 43.7         & 18.7 & -              & -              & \underline{87.6} & \textbf{88.8} \\
                              & BizDocs     & 66.1         & 10.8          & -              & -              & \underline{95.4}          & \textbf{95.9} \\ \midrule
\multirow{2}{*}{\textbf{CLS}} & RVL-CDIP   & 68.2         & 32.8          & -              & -              & \underline{90.9}          & \textbf{91.8} \\
                              & BizDocs     & 84.9         & 40.9          & -              & -              & \underline{98.3}          & \textbf{99.4} \\ \bottomrule \\
\end{tabular}%
}
\end{table}

\begin{table}[]
\caption{\centering Performance comparison on three held-out VRDU datasets in the STDD setting against non-multimodal LLMs.}
\centering
\begin{tabular}{@{}ccc|cccccc@{}}
\toprule
\multirow{2}{*}{\textbf{Model}} &
  \multirow{2}{*}{\textbf{Size} } &
  \multirow{2}{*}{\textbf{Setting}} &
  \textbf{DocVQA} 
  &
  \textbf{KLC} 
  &
   &
  \multicolumn{3}{c}{\textbf{BizDocs}} \\ 
           &          &      &       VQA        &         KIE      &  & VQA           & KIE           & CLS           \\ \midrule
GPT-4+OCR  & $\sim$1T & ZS   & \textbf{82.8}          & \underline{45.9}          &  & \textbf{76.4}          & \underline{66.1}          & \textbf{84.9}          \\
Llama2+OCR & 7B       & ZS   & 47.4          & 27.8          &  & 48.4          & 10.8          & \underline{40.9}          \\ \midrule
DocLLM-1B     & 1B       & STDD & 53.5 & 40.1 & & 65.5  & 63.0 & 20.8\\
DocLLM-7B     & 7B       & STDD & \underline{63.4} &  \textbf{49.9} & & \underline{73.3}  & \textbf{72.6} & 31.1 \\ \bottomrule \\
\end{tabular}

\end{table}

\input{sec_ablation}

\section{Discussion and Findings}
In addition to its immediate utility in visually rich document understanding tasks, we posit that \texttt{DocLLM} offers an opportunity to change the landscape of generative pre-training by enabling language models to go beyond next token prediction in plain text settings. By accommodating complex layout structures, \texttt{DocLLM} allows for e-books, e-publications, and other documents with rich layouts to be incorporated into the pre-training corpus without requiring extensive preprocessing. The spatial-aware reading approach enables the model to perceive the document as inherently structured knowledge. 

Moreover, the multi-page awareness, of both page breaks and document boundaries, enhances the model's ability to comprehend documents of various lengths. This addresses the limitations of previous smaller multi-modal models (which are mainly for single-page documents) and the existing multimodal LLMs (which are primarily designed for images). In supervised instruction tuning, we can adhere to the established practices used in other works, based on desired outputs such as text or images.

The main concept for a cohesive block is to ensure meaningful infilling during the pre-training phase, preventing disconnected predictions. However, the choice of OCR engines to obtain such cohesive blocks remains an open area for exploration. Practical comparisons with various OCR engines and/or layout parsers are left as future work, as LayoutLMs underscore the importance of accurate OCR for improved VQA results. They leverage the Microsoft Azure API, demonstrating superior performance compared to TesseractOCR, as indicated in the DocVQA leaderboard.\footnote{\url{https://rrc.cvc.uab.es/?ch=17&com=evaluation&task=1}} Consequently, researchers are also encouraged to utilize more accurate OCR engines for potential enhancements, if such resources are available.


We have presented a collection of SDDS results alongside zero-shot outcomes. To mitigate prompt influence in the zero-shot results, a rigorous methodology was implemented. This involves the engagement of three independent prompt engineers, each undergoing five rounds of refinement for zero-shot settings, followed by a series of post-processing techniques to enhance result reliability. The best results are thus obtained from each of the three groups. We still acknowledge the potential for refinement and improvement.

We share some internal training experiences, acknowledging the absence of robust validation. First, we observe that a higher weight decay (e.g., 0.1 versus 0.01) generally improves performance in both pre-training and instruction-tuning. During the instruction tuning phase, a higher initial learning rate, such as 1e-4 versus 5e-5, leads to enhanced performance. Overall, we've observed that the cosine scheduler tends to outperform linear or constant schedulers across various settings.

\section{Conclusions}

In this paper, we introduced \texttt{DocLLM}, a lightweight extension to traditional large language models, tailored for generative reasoning over documents with rich layouts. Unlike existing multimodal LLMs, \texttt{DocLLM} strategically omits costly image encoders, instead prioritizing bounding box information to effectively capture the spatial layout structure of documents. This is achieved through a disentangled attention approach, decomposing the attention mechanism in classical transformers, and enhancing with cross-alignment between text and spatial modalities in structured documents. Notably, our model addresses the challenges posed by irregular layouts and heterogeneous content by employing a pre-training objective that focuses on learning to infill block texts. We fine-tuned the pre-trained model using a comprehensive instruction dataset. 
Our evaluation across various document intelligence tasks demonstrates that \texttt{DocLLM} surpasses equivalent models on known tasks for 14 datasets out of 16 and exhibits robust generalization to previously unseen datasets in 4 out of 5 settings, affirming its efficacy in extracting meaningful information from a wide range of visual documents. In future work, we plan to infuse vision into \texttt{DocLLM} in a lightweight manner.



\section*{Acknowledgments}{This paper was prepared for information purposes by the Artificial Intelligence Research group of JPMorgan Chase \& Co and its affiliates (“JP Morgan”), and is not a product of the Research Department of JP Morgan.  J.P. Morgan makes no representation and warranty whatsoever and disclaims all liability for the completeness, accuracy or reliability of the information contained herein. This document is not intended as investment research or investment advice, or a recommendation, offer or solicitation for the purchase or sale of any security, financial instrument, financial product or service, or to be used in any way for evaluating the merits of participating in any transaction, and shall not constitute a solicitation under any jurisdiction or to any person, if such solicitation under such jurisdiction or to such person would be unlawful. © 2023 JP Morgan Chase \& Co. All rights reserved.}

\bibliographystyle{unsrt}  
\bibliography{references}  
\end{document}

%% file: sec_abstract.tex
\begin{abstract}

Enterprise documents such as forms, invoices, receipts, reports, contracts, and other similar records, often carry rich semantics at the intersection of textual and spatial modalities. The visual cues offered by their complex layouts play a crucial role in comprehending these documents effectively. In this paper, we present \docllm, a lightweight extension to traditional large language models (LLMs) for reasoning over visual documents, taking into account both textual semantics and spatial layout. Our model differs from existing multimodal LLMs by avoiding expensive image encoders and focuses exclusively on bounding box information to incorporate the spatial layout structure. Specifically, the cross-alignment between text and spatial modalities is captured by decomposing the attention mechanism in classical transformers to a set of disentangled matrices. Furthermore, we devise a pre-training objective that learns to infill text segments. This approach allows us to address irregular layouts and heterogeneous content frequently encountered in visual documents. The pre-trained model is fine-tuned using a large-scale instruction dataset, covering four core document intelligence tasks. We demonstrate that our solution outperforms SotA LLMs on 14 out of 16 datasets across all tasks, and generalizes well to 4 out of 5 previously unseen datasets.  
\end{abstract}

%% file: sec_intro.tex
\section{Introduction}


Documents with rich layouts, including invoices, receipts, contracts, orders, and forms, constitute a significant portion of enterprise corpora. The automatic interpretation and analysis of these documents offer considerable advantages~\cite{kunduru2023data}, which has spurred the development of AI-driven solutions. These visually rich documents feature complex layouts, bespoke type-setting, and often exhibit variations in templates, formats and quality. Although Document AI (DocAI) has made tremendous progress in various tasks including extraction, classification and question answering, there remains a significant performance gap in real-world applications. In particular, accuracy, reliability, contextual understanding and generalization to previously unseen domains continues to be a challenge~\cite{cui2021document}.

Document intelligence is inherently a multi-modal problem with both the text content and visual layout cues being critical to understanding the documents. It requires solutions distinct from conventional large language models such as GPT-3.5~\cite{brown2020language}, Llama~\cite{touvron2023llama}, Falcon~\cite{penedo2023refinedweb} or PaLM~\cite{anil2023palm} that primarily accept text-only inputs and assume that the documents exhibit simple layouts and uniform formatting, which may not be suitable for handling visual documents. 
Numerous vision-language frameworks~\cite{li2022dit,huang2022layoutlmv3} that can process documents as images and capture the interactions between textual and visual modalities are available. However, these frameworks necessitate the use of complex vision backbone architectures~\cite{dosovitskiy2021an} to encode image information, and they often make use of spatial information as an auxiliary contextual signal~\cite{xu-etal-2021-layoutlmv2, lee-etal-2022-formnet}. 

In this paper we present \texttt{DocLLM}, a light-weight extension to standard LLMs that excels in several visually rich form understanding tasks. Unlike traditional LLMs, it models both spatial layouts and text semantics, and therefore is intrinsically multi-modal. The spatial layout information is incorporated through bounding box coordinates of the text tokens obtained typically using optical character recognition (OCR), and does not rely on any vision encoder component. Consequently, our solution preserves the causal decoder architecture, introduces only a marginal increase in the model size, and has reduced processing times, as it does not rely on a complex vision encoder. We demonstrate that merely including the spatial layout structure is sufficient for various document intelligence tasks such as form understanding, table alignment and visual question answering.

Existing efforts to incorporate spatial layout information typically involve either concatenating  spatial and textual embeddings~\cite{tang2023udop} or summing the two~\cite{xu2020layoutlm}. In contrast, we treat the spatial information as a distinct modality and compute its inter-dependency with the text modality in a disentangled manner~\cite{meng2021connecting}. In detail, we extend the self-attention mechanism of standard transformers to include new attention scores that capture cross-modal relationships. This is motivated by the observation that there is often a correlation between the content, position and size of the fields in a form. Representing their alignments at various abstraction levels across the transformer layers can enhance document understanding.

\begin{figure}
    \centering
\includegraphics[width=\textwidth]{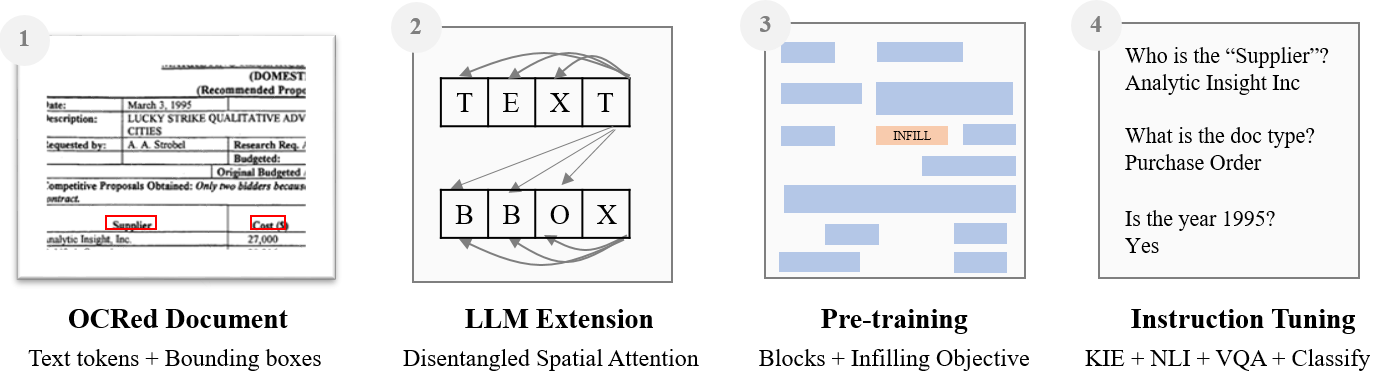}
    \caption{Key elements of \texttt{DocLLM}. (1) Input documents contain text tokens and their bounding boxes. (2) Attention mechanism of LLMs are extended to capture dependencies between text semantics and spatial layouts. (3) Infilling text blocks is used as pre-training objective. (4) Task adaptation is performed on a newly collated dataset of instructions.     }
    \label{fig:overview}
\end{figure}

A common characteristic of visual documents is their heterogeneous content, irregular layouts, and disjointed text segments. When working with such documents, employing a classical next token prediction objective during the self-supervised pre-training phase can be restrictive. In particular, the preceding tokens may not always be relevant due to the diverse arrangements of text, which can be positioned horizontally, vertically, or even in a staggered manner. To tackle this issue, we propose two modifications to the pre-training objective: (a) adopting cohesive blocks of text that account for broader contexts, and (b) implementing an infilling approach by conditioning the prediction on both preceding and succeeding tokens. Due to these modifications, the model is better equipped to address misaligned text, contextual completions, intricate layouts, and mixed data types. Although text spans and infilling tasks have been studied before~\cite{du2021glm}, our solution is tailored for visual documents with an emphasis on semantically coherent blocks.  

We adapt the pre-trained knowledge of \texttt{DocLLM} for several document intelligence tasks by fine-tuning it on instruction data curated from several datasets. These tasks encompass key information extraction, natural language inference, visual question-answering and document classification. Our instruction-tuning data covers both single and multi-page documents.
Layout hints such as field separators, titles and captions can be integrated during instruction-tuning to facilitate learning the logical structure of the documents. We observe that the modifications introduced by \texttt{DocLLM} result in a performance improvement ranging from 15\% to 61\% for the Llama2-7B model in four out of five previously unseen datasets. 

Fig. ~\ref{fig:overview} summarizes the framework. Our contributions include:
\begin{enumerate}
    \item 
    A light-weight extension to LLMs designed for understanding visual documents.
    \item 
    A disentangled spatial attention mechanism that captures cross-alignment between text and layout modalities.
    \item 
    An infilling pre-training objective tailored to address irregular layouts effectively.
    \item 
    An instruction-tuning dataset specially curated towards visual document intelligence tasks.
    \item 
    Comprehensive experiments and valuable insights into the model behavior.
\end{enumerate}

%% file: sec_relatedwork.tex
\section{Related Work}
\label{sec:related_work}


\subsection{LLMs}
The remarkable success of ChatGPT has generated substantial research interest in LLMs across academia and industry. Subsequently, numerous LLMs have been introduced starting from text-based LLMs \cite{openai2023gpt4, chowdhery2022palm,touvron2023llama, workshop2022bloom} to multimodal LLMs \cite{li2023blip,zhu2023minigpt,liu2023visual,wu2023visual,ye2023mplugowl}. In this section, we review these recent advances in LLMs and discuss their connection to and distinctions from our work.   

\textbf{Text-based LLMs}. The introduction of the transformer model in 2017 \cite{vaswani2017attention} has been foundational for the pre-trained models such as BERT \cite{kenton2019bert}, GPT \cite{radford2019language}, and T5 \cite{raffel2020exploring}, each designed with specific pre-training objectives. The emergence of ChatGPT and GPT-4 marked a notable shift, characterized by a substantial increase in both model parameters and training data size. This enhancement has resulted in remarkable zero-shot generalization capabilities, allowing these models to excel in tasks previously unseen. Such success of LLMs has prompted the development of additional LLMs such as OPT \cite{zhang2019root}, BLOOM \cite{workshop2022bloom}, PaLM \cite{chowdhery2022palm}, and Llama \cite{touvron2023llama}.
Particularly, Llama2 \cite{touvron2023llama} is an open-source LLM that achieves comparable or better performance to both open and closed-sourced models, including ChatGPT, PaLM and Falcon, with enhanced safety strategies. Llama2 employs the standard Transformer architecture with pre-normalization \cite{zhang2019root}, SwiGLU activation function \cite{shazeer2020glu}, and rotary positional embeddings \cite{su2023roformer}. The pre-training data consists of two trillion tokens from publicly available sources.

\textbf{Multimodal LLMs}. Multimodal LLMs extend the scope of text to diverse modalities, with a focus on visual input. These models can be categorized into two tropes: general-purpose multimodal LLMs \cite{li2023blip,zhu2023minigpt,liu2023visual,wu2023visual,ye2023mplugowl} and models that are tailored for visually-rich document understanding \cite{ye2023mplugdocowl,ye2023ureader,kim2022donut,lee2023pix2struct,tang2023udop}. The general-purpose multimodal LLMs exhibit promising performance in identifying and reasoning with image information. However, they have not yet been vigorously evaluated on VRDU tasks. As an example, the GPT-4 Technical Report \cite{openai2023gpt4} highlights diverse multimodal test cases, such as explaining meme picture distinctiveness, but very few examples are included for visual document use cases. Prior to the advent of large language models, fine-tune-based models relying on vision only were less effective than layout (and vision) modality models in processing visual documents. For example, models like UDOP \cite{tang2023udop} and LayoutLM \cite{xu2020layoutlm} outperform vision-only models such as Donut \cite{kim2022ocrfree} and Pix2Struct \cite{lee2023pix2struct} in VRDU tasks. But such models require task- and dataset-specific fine-tuning, and are thus excluded in our analysis. The more recent mPLUG-DocOwl \cite{ye2023mplugdocowl} and UReader \cite{ye2023ureader}, built upon LLMs, undergo instruction finetuning on a diverse set of VRDU, visual, and textual datasets, and exhibit impressive zero-shot generalization capabilities. Hence, we include those as baselines in our evaluation in Section \ref{sec:experiments}.

Despite the remarkable performance of LLMs, unimodal models aren't equipped to process multimodal input, and multimodal LLMs rely on complex and memory intensive open-domain vision encoders. Our proposed model, \docllm, addresses these challenges by explicitly modeling spatial layouts and text semantics, enabling effective comprehension of visual documents. Notably, \docllm offers an extension to the unimodal architecture by adding the spatial signal to text semantics, avoiding the expensive vision encoder, resulting in a more compact model and efficient processing time.

\subsection{LLM Architectures}
\label{sc:auto_infill}
\textbf{Autoregressive Infilling}. 
There are two main autoregressive infilling approaches: ``fill-in-the-middle'' (FIM) where a single span is sampled, and ``blank infilling'' with multiple spans. 

The OpenAI FIM approach \cite{bavarian2022efficient} uses the template \texttt{(prefix, middle, suffix)} to divide a document into three segments. Next, these segments are reorganized into \texttt{(prefix, suffix, middle)}, enabling the model to predict the middle segment. This process relies on three special tokens, \texttt{[PRE]}, \texttt{[SUF]}, and \texttt{[MID]}, which structure a document as: \texttt{[PRE]} prefix \texttt{[SUF]} suffix \texttt{[MID]} middle. The \texttt{[MID]} token denotes the start for prediction, while the other two special tokens guide the model on where to infill. This method demonstrates that autoregressive models can learn to infill text where the middle part is missing. Fill-in Language Model (FiLM) \cite{shen2023film} is a subsequent development that enables flexible generation at arbitrary positions, unconstrained by a predefined generation order. In contrast, approaches like GLM \cite{du2021glm} sample multiple spans for infilling. For each blank to be infilled, a pair of special tokens is used: \texttt{[blank\_mask]} and \texttt{[start\_to\_fill]}. The multiple spans not only require special tokens but also global indicators to distinguish which middle span the model should infill. This global indicator is implemented with 1D token positions, ensuring that each pair of the two special tokens, i.e., \texttt{[blank\_mask]} and \texttt{[start\_to\_fill]}, share the same positions.
We adopt a similar infilling object with the goal to prevent disconnected next-token predictions while avoiding breaking sparse documents into very short segments, e.g., word pieces and/or phrase pieces. 

\textbf{Disentangled attention}. Disentangled attention is introduced in the DeBERTa model \cite{he2020deberta}, where token embeddings and relative positional encodings were kept separate rather than summed together, and each used independently when computing attention weights using disentangled matrices. The motivation behind this was to facilitate the learning of decoupled attention alignments based on content and position separately. This innovation proved effective as it allowed DeBERTa to outperform RoBERTA-large and T5 on NLU benchmarks, as well as to surpass the human baseline on SuperGLUE \cite{wang2019superglue}. In our work, given considerably more complex position encodings used in visually rich documents, disentanglement becomes ever more important to our model's performance.

%% file: sec_pretrain.tex

\begin{figure}
\centering
\includegraphics[width=\textwidth]{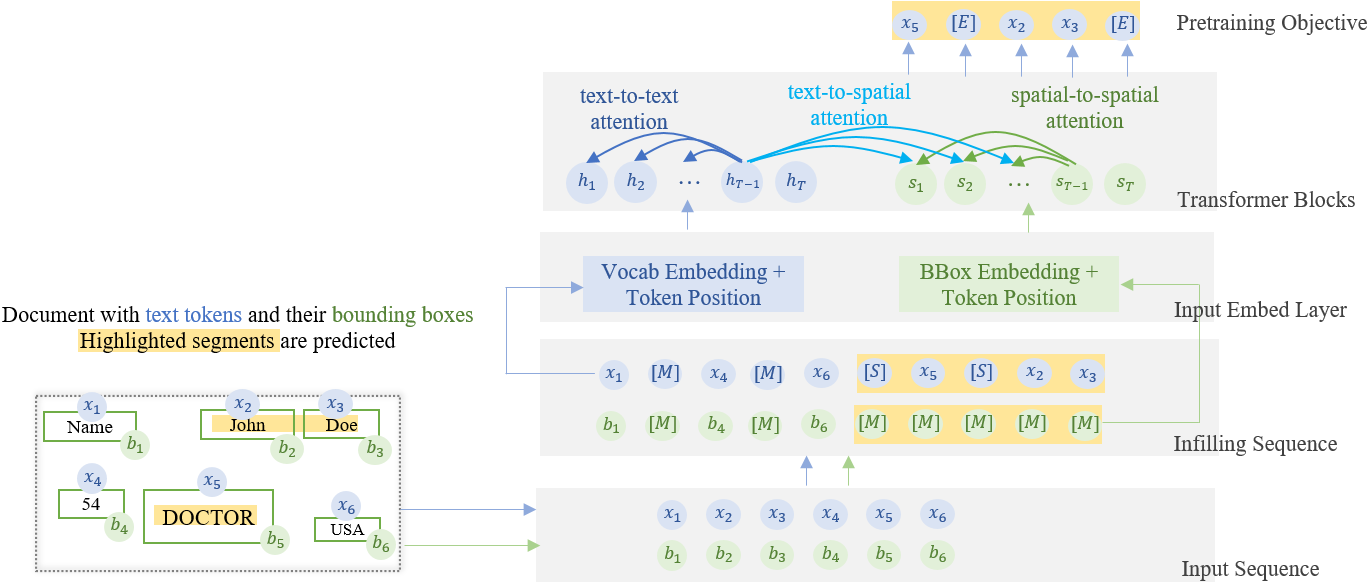}
\caption{\docllm model architecture with disentangled spatial attention and infilling objective. \emph{left}: Input document with text tokens $x_i$ and bounding boxes $b_i$. Some text segments are randomly masked (two segments here) and the model predicts the tokens in these text segments autoregressively. \emph{right}: The infilling sequence is created by replacing the sampled segments with $\mathrm{[M]}$ and prepending them with $\mathrm{[S]}$. The attention mechanism is extended to account for cross-attention between text and spatial modalities.} 
\label{fig:modelarch}
\end{figure}

\subsection {Model Architecture}
\texttt{DocLLM} is constructed upon the foundation of an auto-regressive transformer language model \cite{touvron2023llama} following a causal decoder structure. It is composed of stacked transformer blocks, where each block contains a multi-head self-attention layer and a fully connected feed forward network. Standard language models are typically unimodal, accepting only a sequence of text tokens as input. In contrast, \texttt{DocLLM} is a multi-modal system that integrates lightweight visual information by utilizing the spatial positions and dimensions of text tokens obtained using OCR. Simply augmenting the text with bounding box information via additive positional encoding may not capture the intricate relationships between text semantics and spatial layout, especially for visually rich documents \cite{xu-etal-2021-layoutlmv2}. Consequently, we treat the spatial information about the text tokens as a distinct modality. In particular, we use separate vectors to represent these two modalities and extend the self-attention mechanism of the transformer architecture to compute their inter-dependencies in a disentangled manner, as explained in the following section. Furthermore, instead of the traditional left-to-right next token prediction during self-supervised training, we employ a text infilling objective that better leverages contextual information.

\subsection {Disentangled Spatial Attention}
Let $\mathbf{x}=(x_1,...,x_i,...,x_T)$ be an input sequence of length $T$, where $x_i$ is a text token. In classical transformers, using a learned embedding matrix based on the text vocabulary and a learned set of parameters for the token position in the sequence, the input tokens are first encoded into hidden vectors $\mathbf{H} \in \mathbb{R}^{T \times d}$. A self-attention head then computes the attention scores between tokens $i$ and $j$ as:
\begin{align}
\mathbf{Q}^t &= \mathbf{H}\mathbf{W}^{t, q}, &  \mathbf{K}^t  &= \mathbf{H}\mathbf{W}^{t, k}, & \mathbf{A}^t_{i,j} &=\mathbf{Q}^t_i{\mathbf{K}^t_j}^\intercal 
\end{align}
where  $\mathbf{W}^q \in \mathbb{R}^{d \times d}$ and $\mathbf{W}^k \in \mathbb{R}^{d \times d}$ are projection matrices, and the superscript $t$ indicates the text modality. The attention scores $\mathbf{A} \in \mathbb{R}^{T \times T}$ along with another projection matrix $\mathbf{W}^v$ are further used to compute the hidden vectors $\mathbf{H'}$, which are in turn used as inputs for a subsequent layer:
\begin{align}\label{eqn:nlayer}
\mathbf{V}^t &= \mathbf{H}\mathbf{W}^{t, v}, &  \mathbf{H'} &= softmax(\frac{\mathbf{A}^t}{\sqrt{d}})\mathbf{V}^t. 
\end{align}

In \texttt{DocLLM}, the input is represented as $\mathbf{x}=\{(x_i,b_i)\}_{i=1}^T$, where $b_i=\texttt{(left, top, right, bottom)}$ is the bounding box corresponding to $x_i$. To capture the new modality (i.e. spatial information), we encode the bounding boxes into hidden vectors represented by $\mathbf{S} \in \mathbb{R}^{T \times d}$.
We then decompose the attention matrix computation into four different scores, namely \emph{text-to-text}, \emph{text-to-spatial}, \emph{spatial-to-text} and \emph{spatial-to-spatial}. Formally, the new attention mechanism is calculated as:
\begin{align}
\mathbf{Q}^s &= \mathbf{S}\mathbf{W}^{s,q}, &  \mathbf{K}^s  &= \mathbf{S}\mathbf{W}^{s,k} \\
\mathbf{A}_{i,j} &=\mathbf{Q}^t_i{\mathbf{K}^t_j}^\intercal +
\lambda_{t,s}\mathbf{Q}^t_i{\mathbf{K}^s_j}^\intercal + \lambda_{s,t}\mathbf{Q}^s_i{\mathbf{K}^t_j}^\intercal + 
\lambda_{s,s}\mathbf{Q}^s_i{\mathbf{K}^s_j}^\intercal,
\end{align}

where $\mathbf{W}^{s,q} \in \mathbb{R}^{d \times d}$ and $\mathbf{W}^{s,k} \in \mathbb{R}^{d \times d}$ are newly introduced projection matrices corresponding to the spatial modality, and $\lambda$s are hyperparameters that control the relative importance of each score. The input hidden vectors for the next layer $\mathbf{H'}$ are computed exactly as before. However, in contrast to equation (\ref{eqn:nlayer}), the newly calculated hidden vectors rely not only on the text semantics but also on the layout information of the text tokens.

It is important to mention that the hidden vectors $\mathbf{S}$ are reused across different layers, while each layer retains the flexibility to employ different projection matrices. We also note that the number of extra parameters required to encode the bounding box information is significantly lower compared to the overhead introduced by image based models \cite{li2022dit}. By simply adding $\mathbf{S}$ to $\mathbf{H}$ similar to ~\cite{xu2020layoutlm}, we could have avoided using $\mathbf{W}^s$ matrices altogether and further reduced the number of parameters. However, it would have irreversibly  coupled the layout information with the text semantics. In contrast, our disentangled representation of these modalities in the attention scores enables selective focus when appropriate \cite{he2020deberta}, thereby providing an optimal balance between model size and effectiveness.

\subsection{Pretraining}
\texttt{DocLLM} is first pre-trained in a self-supervised fashion on a large number of unlabeled documents. The self-supervised pre-training objective in autoregressive language models \cite{radford2019language} is generally to maximize the log-likelihood of the next token prediction in a sequence based on the context provided by preceding tokens. Let $\theta$ denote all the parameters of the transformer model, including the projection matrices discussed above. The following cross-entropy loss is then typically minimized during the pre-training step: 
\begin{align}
\mathcal{L}_{\text{AR}}(\theta) = -\sum_{i=1}^T\log p_{\theta}(x_i | \mathbf{x}_{j<i})
\end{align}

 Visual documents are often sparse and irregular, featuring isolated and disconnected text fragments. In such cases, it is preferable to consider coarse segments of related tokens during pre-training rather than focusing on individual tokens. A segment may represent a coherent chunk of information, similar to a text block, or it can simply be a linear sequence, similar to a text span. In Figure \ref{fig:modelarch}, ``Name'', ``John Doe'' , and ``Doctor'' are all examples of blocks. In general, the broader context provided by multiple tokens in a block can lead to better comprehension. 
 
 Furthermore, learning to infill text, where the prediction is conditioned on both prefix and suffix tokens rather than only preceding tokens, can be beneficial. The infilling objectives enable contextually relevant completions, provide robustness to OCR noise or misaligned tokens, and can better handle relationships between various document fields. Hence we modify the standard pre-training objective to predict blocks of text given preceding and following text blocks.

 Most OCR engines can provide block level information, which makes it feasible to identify coherent text blocks such as a heading or an address\footnote{Note that in order to avoid any leakage of useful information, the block information is only used for the masking objective during pre-training, and is not provided to the model as input. Concretely, masking is performed at the block level, but the model is not provided with information about the number of tokens in a given masked block. Please refer to Figure \ref{fig:modelarch} for an illustrated example.}. 
 Inspired by ~\cite{du2021glm}, we follow an autoregressive block infilling objective, where text blocks are randomly masked, and the masked blocks are shuffled and reconstructed in a sequential left-to-right fashion. Block information and block infilling are solely utilized for the pre-training phase, not in instruct-tuning or downstream tasks. 
 
 Formally, let $\mathbf{c}=\{c_1,...,c_K\}$ be a set of text blocks that partitions an input sequence $\mathbf{x}$ into non-overlapping contiguous tokens such that $c_1 \cup ... \cup c_K = \mathbf{x}$ and $c_k \cap c_{k'} = \emptyset$. These text blocks are typically identified from OCR information. Let $\mathbf{z}=\{z_m\}_{m=1}^M$ be $M\ll K$ different text blocks randomly sampled from $\mathbf{c}$, where each block $z_m=(z_{m,1},...,z_{m,N_m})$ contains a consecutive series of tokens. Further, let $\mathbf{\tilde{x}}$ be a corrupted version of $\mathbf{x}$ where the contiguous tokens corresponding to a sampled text block are replaced with a special mask token $\mathrm{[M]}$. To facilitate the identification of the block to be filled during text generation, each input block is augmented with a special  start token $\mathrm{[S]}$ while the output block includes an end token $\mathrm{[E]}$. For instance, a block with tokens $(x_4, x_5)$ becomes $\mathrm{[M]}$ in $\mathbf{\tilde{x}}$, $(\mathrm{[S]}, x_4, x_5)$ when conditioned upon, and is expected to generate $(x_4, x_5, \mathrm{[E]})$ as output autoregressively (see Figure \ref{fig:modelarch} for a detailed illustration of these configurations). The following cross-entropy loss is then minimized for the infilling objective. 
 
 \begin{align}
\mathcal{L}_{\text{IF}}(\theta) = -\sum_{m=1}^M \sum_{j=1}^{N_m} \log p_{\theta}(z_{m,j} | \mathbf{\tilde{x}},  \mathbf{z}_{<m}, \mathbf{z}_{m,<j})
\label{eq:disent_att}
\end{align}

%% file: sec_instructtune.tex
\subsection{Instruction Tuning}
\label{sec:instruction_tuning}

\begin{table}[]
\caption{\label{templates} Prompt templates used for instruction-tuning (spatial tokens not included).}
\centering
\resizebox{\columnwidth}{!}{%
\begin{tabular}{@{}llll@{}}
\toprule
\textbf{Task} &
  \textbf{Template type} &
  \textbf{Prompt template} &
  \textbf{Expected response} \\ \midrule
\textbf{VQA} &
  Extraction &
  "\{\texttt{document}\} \{\texttt{question}\}" &
  \texttt{answer} annotation \\ \midrule
\textbf{NLI} &
  MCQ &
  "\{\texttt{document}\} \textbackslash{}"\{\texttt{statement}\}\textbackslash{}", Yes or No?" &
  \texttt{answer} annotation \\ \midrule
\multirow{3}{*}[-1em]{\textbf{KIE}} &
  Extraction &
  "\{\texttt{document}\} What is the value for the \textbackslash{}"\{\texttt{key}\}\textbackslash{}"?" &
  Associated \texttt{value} annotation \\ \cmidrule{2-4}
 &
  MCQ &
  \begin{tabular}[c]{@{}l@{}}"\{\texttt{document}\} What is \textbackslash{}"\{\texttt{value}\}\textbackslash{}" in the document? Possible choices: \{\texttt{choices}\}."\\ \footnotesize \textit{(where {\normalfont \texttt{choices}} is a subset of all the keys in the dataset in random order)}\end{tabular} &
  Associated \texttt{key} annotation \\ \cmidrule{2-4}
 &
  Internal classification &
  "\{\texttt{document}\} What is \textbackslash{}"\{\texttt{value}\}\textbackslash{}" in the document?" &
  Associated \texttt{key} annotation \\ \midrule
\multirow{2}{*}[-0.25em]{\textbf{CLS}} &
  MCQ &
    \begin{tabular}[c]{@{}l@{}}"\{\texttt{document}\} What type of document is this? Possible choices: \{\texttt{choices}\}."\\ \footnotesize \textit{(where {\normalfont \texttt{choices}} is a subset of all the classes in the dataset in random order)}\end{tabular} &
  \texttt{class} annotation \\ \cmidrule{2-4}
 &
  Internal classification &
  "\{\texttt{document}\} What type of document is this?" &
  \texttt{class} annotation \\ \bottomrule
\end{tabular}%
}
\end{table}


Following recent work in the field of VRDU \cite{tang2023udop,ye2023mplugdocowl,ye2023ureader}
 and prior work in NLP \cite{wei2022flan1,chung2022flan2}, we instruction-tune \texttt{DocLLM} on a variety of instructions derived from DocAI datasets using 
various templates. Due to the high cost and time intensity of manual data collection, we leave the construction of a VRDU instruction-tuning dataset with crowdsourced instructions and preferences to future work. We employ a total of 16 datasets with their corresponding OCRs, spanning four DocAI tasks:  visual question answering (VQA), natural language inference (NLI), key information extraction (KIE), and document classification (CLS). 
 
The diversity of supervised fine tuning (SFT) instructions is critical in helping zero-shot generalization \cite{wei2022flan1,chung2022flan2,ouyang2022instructgpt}. Thus, we diversify templates per task when possible, with each template asking a different question, and in some cases, expecting different types of answers. We re-use the templates introduced in \cite{ye2023mplugdocowl,ye2023ureader} when applicable, and consider a broader selection of datasets in our instruction-tuning data mix. 

We create the templates following what we believe end users would generally ask about documents (Table~\ref{templates}). For KIE and CLS, we hypothesize that (1) the extraction instructions can teach \texttt{DocLLM} to correlate names of keys in the prompts with document fields so as to retrieve values, (2) the internal classification instructions can help the model understand what intrinsically characterizes each key or document type, and (3) the multiple choice question (MCQ) instructions can teach the model to leverage its comprehension of key names included as choices in the prompt (resp. document type names) to classify extracted values (resp. entire documents).
We introduce the templates in detail as follows. 

     \textbf{Visual Question Answering}. We collect DocVQA \cite{mathew2021docvqa}, WikiTableQuestions (WTQ) \cite{pasupat2015wtq}, 
     VisualMRC \cite{tanaka2021visualmrc}, DUDE \cite{landeghem2023dude}, and BizDocs\footnote{BizDocs is a collection of business entity filings that is due to be released publicly.}, to compose the VQA instruction-tuning data mix. We use one instruction template to build our SFT inputs for VQA, as shown in table \ref{templates}. An example prompt derived from DocVQA would read: "\{\texttt{document}\} \textit{What is the deadline for scientific abstract submission for ACOG - 51st annual clinical meeting?}"
     
     \textbf{Natural Language Inference}. We only include TabFact \cite{chen2020tabfact} in our instruction-tuning data mix for NLI task, due to lack of additional DocAI NLI datasets available. The instruction template is shown in table \ref{templates}. An example prompt derived from TabFact would read: "\{\texttt{document}\} \textit{\textbackslash"The UN commission on Korea include 2 Australians.\textbackslash", Yes or No?}"
     
    \textbf{Key Information Extraction}. We gather Kleister Charity (KLC) \cite{stanislawek2021klc}, CORD \cite{park2019cord}, FUNSD \cite{jaume2019funsd}, DeepForm \cite{svet2020deepform}, PWC \cite{kardas2020axcell}, SROIE \cite{huang2019sroie}, VRDU ad-buy 
    \cite{wang2023vrdu} (with random train-test splitting), and BizDocs to build the KIE instruction-tuning data, where we leverage three instruction templates: extraction, internal classification, and MCQ, as shown in \ref{templates}.
    For the extraction template, we add the ``None'' answer if the key does not exist in the given document. To increase diversity in the SFT training data, we also derive internal classification and MCQ instructions from original KIE annotations. To stay consistent with benchmarks from previous work \cite{ye2023mplugdocowl,ye2023ureader}, we only keep the prompts derived from the extraction template in the test split of each KIE dataset. An example extraction instruction derived from KLC would read: "\{\texttt{document}\} \textit{What is the value for the \textbackslash"charity number\textbackslash"?}"
    
    \textbf{Document Classification}. We aggregate RVL-CDIP \cite{harley2015rvlcdip} and BizDocs to build our CLS instruction-tuning data. 
    We used two types of instruction templates for this task: internal classification and MCQ, as shown in \ref{templates}.
    To avoid the cold start problem induced by potentially unseen types of documents in testing or even in production usage, we only keep the MCQ prompts for the test split of each CLS dataset.  We also downsample RVL-CDIP in the train split to avoid hindering the other datasets. An example MCQ instruction derived from RVL-CDIP would read: "\{\texttt{document}\} \textit{What type of document is this? Possible answers: [budget, form, file folder, questionnaire].}"


%% file: sec_ablation.tex
\section{Ablation Studies}




We conduct ablation studies to validate the three contributions of \texttt{DocLLM}: (1) disentangled spatial features, (2) the block infilling pre-training objective, and (3) the masking strategy used for decoding. 

For all ablations, we use Next Token Prediction (NTP) out-of-sample accuracy to compare configurations at the pre-training stage. Due to resource restrictions, each experiment uses a subset of our pre-training corpus: we randomly sample 100,000 chunks and predict on 1,000 unseen documents. A chunk is a pack of documents concatenated one by one with the total length less than maximum input length. The hyperparameters are set consistently following Table \ref{tb:key_settings} across all ablation experiments.

\begin{figure}
\small
  \begin{subfigure}{0.46\textwidth} 
    \centering
    \includegraphics[width=\linewidth]{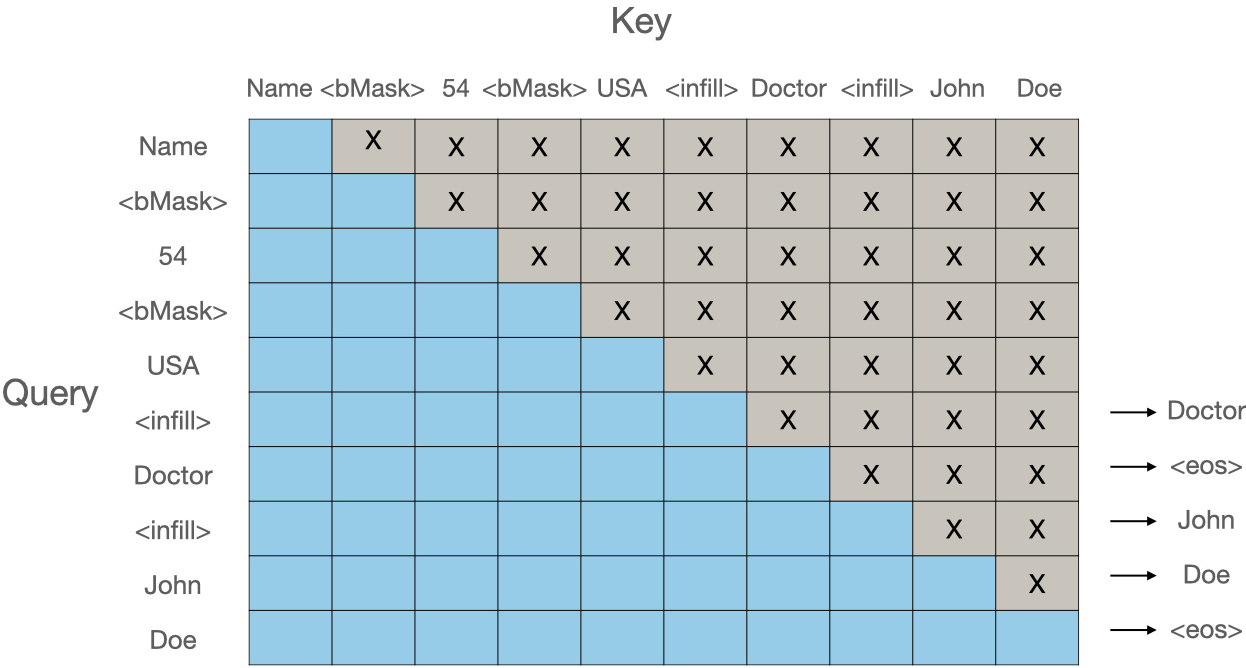}
    \caption{Causal decoder}
    \label{fig:subfig1}
  \end{subfigure}\hfil
  \begin{subfigure}{0.46\textwidth} 
    \centering
    \includegraphics[width=\linewidth]{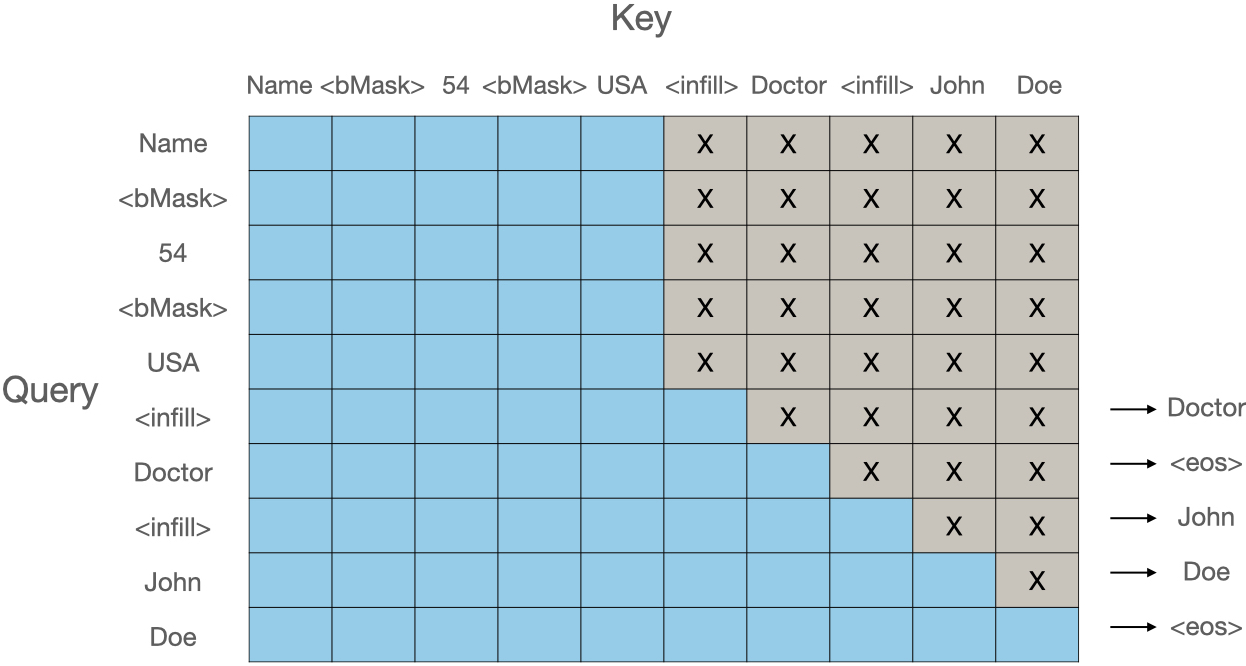}
    \caption{Prefix decoder}
    \label{fig:subfig2}
  \end{subfigure}
  \caption{A simplified illustration of attention masks for causal-decoder and prefix-decoder for block infilling.}
  \label{fig:causauvsprefix}
\end{figure}




\begin{table}[]
\centering
\caption{\centering Ablation study on disentangled spatial attention. \texttt{T} stands for the text modality, \texttt{S} stands for the spatial modality, and their cross-modal interactions represent as \texttt{X2X}, e.g., text-to-spatial $\rightarrow$ \texttt{T2S}.}
\begin{tabular}{@{}rc@{}}
\toprule
\textbf{Cross-Modal Interactions} & \textbf{NTP Accuracy} \\ \midrule
\texttt{T2T}           & 35.43                 \\
\texttt{T2S + T2T}                               & 38.08                 \\
\texttt{S2T + T2T}                               & 38.05                 \\
\texttt{S2S + T2T}                               & \textbf{39.12}                \\
\texttt{T2S + S2S + T2T}                           & \underline{39.06}                 \\
\texttt{S2T + S2S + T2T}                           & \underline{39.07}                 \\
\texttt{T2S + S2T + S2S + T2T}                   & 39.02                 \\ \bottomrule
\end{tabular}
\label{tab:spatial_ablation}
\end{table}


\textbf{Disentangled Spatial Attention}. To measure the effect of disentangled spatial attention on cross-modal interactions, we train the models by setting the $\lambda$ hyperparameter in Eq~\ref{eq:disent_att} to $0$ or $1$ . Table \ref{tab:spatial_ablation} enumerates the attention combinations,  and the results suggest that keeping only the spatial-to-spatial interaction (i.e. $\lambda_{s,s}=1$) yields the highest NTP accuracy. The performance differences among other configurations, such as text-to-spatial and spatial-to-text, are  subtle. Notably, the vanilla text-only self-attention mechanism yields the lowest NTP accuracy, underlining the importance of incorporating spatial features for understanding documents with rich layouts. For all experiments in Section \ref{sec:experiments}, we therefore set $\lambda_{s,s}=1$, $\lambda_{s,t}=0$, and $\lambda_{t,s}=0$. We opt for simplicity by choosing a hard mode over a soft one while acknowledging the potential advantage of flexibility for the latter.

\textbf{Autoregressive Block Infilling}. To evaluate the effectiveness of the proposed autoregressive block infilling objective especially comparing with the conventional left-to-right causal learning, we benchmark three configurations in our ablation study: (1) causal learning, (2) causal learning with spatial modality, and (3) block infilling with spatial modality. As highlighted in Table \ref{tab:obj_ablation}, autoregressive block infilling exhibits the best performance. Additionally, the performance gain of adding the spatial modality to the causal learning proves the advantage of the spatial modality.


\begin{table}[]
\centering
\caption{\centering Ablation study on the block infilling objective.  }
\begin{tabular}{@{}lc@{}}
\toprule
\textbf{Pretraining Objective}      & \textbf{NTP Accuracy} \\ \midrule
Causal Learning                     & 32.6                  \\
Causal Learning + Spatial  & \underline{36.2}                  \\
Block Infilling + Spatial  & \textbf{39.1}                  \\ \bottomrule
\end{tabular}
\label{tab:obj_ablation}
\end{table}


\begin{figure}
    \centering
\includegraphics[width=0.45\textwidth]{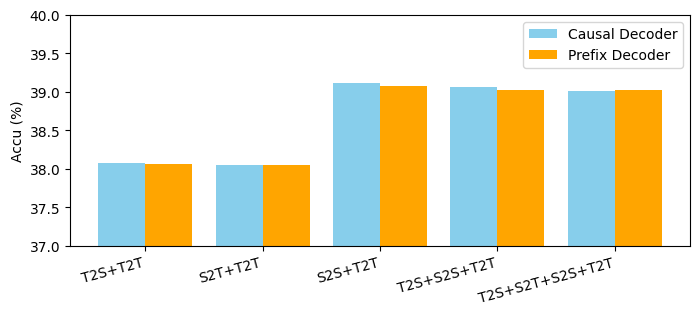}
    \caption{\label{fig:prefix_abl} Performance comparison on NTP between causal decoder and prefix decoder.}
    \label{fig:causauvsprefix_res}
\end{figure}



\textbf{Prefix Decoder and Causal Decoder}. For document-conditioned generation, an intuitive choice is to employ a prefix decoder with prefix masking to make the whole  document bidirectional visible in the attention, as illustrated in Figure \ref{fig:subfig2}. We investigate this assumption through experiments where we compare a prefix decoder against the conventional causal decoder. Specifically, we conduct contrast experiments on these two decoders for different settings outlined in the \textbf{disentangled spatial attention} to study their resulting performance. 

The results in Figure \ref{fig:causauvsprefix_res} show marginal differences between these two decoder across the five configurations, with the causal decoder having a slight edge over the prefix. The minor difference suggests that both masking methods are comparable in modeling documents. 
Thus the bidirectional attention enabled by the prefix decoder may not be crucial in this context, and we consequently elect to use a causal decoder for all experiments in section \ref{sec:experiments}.
